\pdfoutput=1

\documentclass[11pt]{article}

\usepackage[]{acl}

\usepackage{times}
\usepackage{latexsym}
\usepackage{multicol}
\usepackage{multirow}
\usepackage{graphicx}
\usepackage{makecell} 

\usepackage[T1]{fontenc}

\usepackage[utf8]{inputenc}

\usepackage{microtype}

\usepackage{inconsolata}

\title{Pre-Finetuning with Impact Duration Awareness for Stock Movement Prediction}

\author{Chr-Jr Chiu,\textsuperscript{\rm 1}
    Chung-Chi Chen,\textsuperscript{\rm 2}
    Hen-Hsen Huang,\textsuperscript{\rm 3}
    Hsin-Hsi Chen\textsuperscript{\rm 1} \\
   \textsuperscript{\rm 1} Department of Computer Science and Information Engineering, \\ 
   National Taiwan University, Taiwan \\
    \textsuperscript{\rm 2} AIST, Japan \\
    \textsuperscript{\rm 3}
    Institute of Information Science, Academia Sinica, Taiwan \\
    ccchiu@nlg.csie.ntu.edu.tw,
    c.c.chen@acm.org,\\
    hhhuang@iis.sinica.edu.tw,
    hhchen@ntu.edu.tw}

\begin{document}
\maketitle
\begin{abstract}
Understanding the duration of news events' impact on the stock market is crucial for effective time-series forecasting, yet this facet is largely overlooked in current research. This paper addresses this research gap by introducing a novel dataset, the Impact Duration Estimation Dataset (IDED), specifically designed to estimate impact duration based on investor opinions. Our research establishes that pre-finetuning language models with IDED can enhance performance in text-based stock movement predictions. In addition, we juxtapose our proposed pre-finetuning task with sentiment analysis pre-finetuning, further affirming the significance of learning impact duration. Our findings highlight the promise of this novel research direction in stock movement prediction, offering a new avenue for financial forecasting. We also provide the IDED and pre-finetuned language models under the CC BY-NC-SA 4.0 license for academic use, fostering further exploration in this field.
\end{abstract}

\section{Introduction}
\label{sec:introduction}

The arena of text-based financial market prediction has gained significant momentum within the research community, particularly those vested in the intersection of natural language processing (NLP) and finance~\cite{xu-cohen-2018-stock,hu2018listening,qin-yang-2019-say,sawhney-etal-2020-voltage,sawhney-etal-2022-cryptocurrency}. News articles, as a chief vehicle of up-to-the-minute information drawn from a multitude of sources, serve as an influential corpus in this field of study.
Despite the rich collection of research delving into the prediction of market information using news data, there remains a conspicuous gap in understanding the temporal impact of news events on the financial market. We posit that this temporal information bears significant relevance to time-series forecasting. For instance, news related to broader economic trends might exert a more protracted influence than a single news item disclosing a company's monthly earnings. 


The complexity of the financial market is such that consensus is elusive, even among experts. When presented with the same piece of news, investors and professional analysts may interpret it divergently, their perceptions colored by differing analytical perspectives and the range of information at their disposal. An investor's viewpoint typically comprises two components: the news (premise) and the claim~\cite{chen2021evaluating}. Based on the given news, investors project a claim about future market developments.
Table~\ref{tab:Example of different impact duration estimations based on the same news} furnishes an example where distinct professional analysts draw disparate conclusions about the same news, with varying estimates of its impact duration within the same day. Claim A is focused on the revenue for \textbf{2022}, while Claim B anticipates the revenue's contribution to \textbf{the first quarter of 2023}. This highlights the inherent challenges in annotating news articles directly. 

\begin{table}[t]
  \centering
    \resizebox{0.8\columnwidth}{!}{

    \begin{tabular}{l|p{17em}}
    \Xhline{2\arrayrulewidth}
    \multirow{3}[0]{*}{News} & 2022 first quarter's profit hits record high, boosted by the expansion of the semiconductor industry. \\
    \hline
    \multirow{3}[0]{*}{Claim A} & It is estimated that the revenue in \textbf{2022} will increase to 6.019 billion yuan, an increase of 56.99\% YOY. \\
    \hline
    \multirow{5}[0]{*}{Claim B} & The actual start of production may be delayed until the end of the year, and the timing of revenue contribution is also estimated to be in \textbf{the first quarter of 2023}. \\
    \Xhline{2\arrayrulewidth}
    \end{tabular}%
    }
        \caption{Example of different impact duration estimations based on the same news (event).}
  \label{tab:Example of different impact duration estimations based on the same news}%
\end{table}%

As a solution to this problem, we present an innovative approach in this paper: learning from the estimations of investors' opinions. We opt to annotate the impact duration estimates on investors' opinions rather than making annotations on news articles directly. This implies that annotators are tasked with reading and extracting impact duration estimates from the opinions, thereby circumventing the need for personal inferences from annotators.

To test our approach, we conduct experiments on the news-based stock movement prediction benchmark dataset, Astock~\cite{zou2022astock}. Our results demonstrate that pre-finetuning with our proposed Impact Duration Estimation Dataset (IDED) consistently enhances the performance of stock movement prediction, independent of the language models deployed. We further examine other pre-finetuning strategies using sentiment analysis labels, with outcomes indicating that tangible improvements are realized only when integrating our proposed task into the pre-finetuning process.

To sum up, this paper pioneers a new dataset designed for a novel pre-finetuning task, and underscores the potential of this research direction for the stock movement prediction task. To facilitate further academic exploration, we intend to release the proposed IDED along with pre-finetuned language models under the CC BY-NC-SA 4.0 license.

\begin{table}[t]
  \centering
    \resizebox{0.8\columnwidth}{!}{
    \begin{tabular}{l|r}
    \Xhline{2\arrayrulewidth}
     \multicolumn{1}{c|}{Label} &  \multicolumn{1}{c}{Number of Instances} \\
     \hline
    Longer than 1 week & 6,108 \\
    Within 1 week & 1,947 \\
    Unsure & 705 \\
    \Xhline{2\arrayrulewidth}
    \end{tabular}%
    }
      \caption{Statistics of the proposed IDED.}
  \label{tab:Statistics of the proposed IDED}%
\end{table}%

\section{Related Work}
\label{sec:related_work}


There is a rich body of work employing unstructured data for prediction, such as leveraging news articles~\cite{hu2018listening}, transcripts from company meetings~\cite{qin-yang-2019-say,sawhney-etal-2020-voltage}, and social media posts~\cite{xu-cohen-2018-stock,sawhney-etal-2022-cryptocurrency}. These methods aim to capture the sentiment or public opinion that may influence stock movement.
Hybrid models have also been proposed to amalgamate structured and unstructured data, striving to benefit from the strengths of both approaches~\cite{xu-cohen-2018-stock,qin-yang-2019-say,ni2021hybrid}. These models represent an attempt to provide a more holistic perspective by combining the precision of structured data with the rich contextuality of unstructured data.
This paper presents the first dataset dedicated to impact duration estimation and initiating scholarly discussion on this crucial aspect. 
In the realm of language modeling, the pre-finetuning scheme has emerged as a widely accepted method to boost performance in specific tasks~\cite{aghajanyan2021muppet}. This scheme modifies embeddings by learning tasks closely related to the target task. The key challenge in this approach lies in identifying suitable intermediate tasks for training.


\section{Dataset}
\label{sec:dataset}
\subsection{Impact Duration Estimation Dataset}
Our dataset, IDED, comprises investors' opinions harvested from Mobile01,\footnote{\url{https://www.mobile01.com/topiclist.php?f=793}} a widely-used platform for investment discussions. A professional employed at a securities company annotates 8,760 posts, determining the impact duration (within 1 week/longer than 1 week/unsure) of each post. 

To ensure the validity of our annotations, we select a random subset of 1,000 instances and have them annotated by a different expert from the same domain. We evaluate the inter-rater reliability using the Cohen-Kappa statistic~\cite{mchugh2012interrater}, achieving a substantial agreement of 66.49\%~\cite{landis1977measurement}. The distribution of these labels is documented in Table~\ref{tab:Statistics of the proposed IDED}.

\subsection{AStock}

Our experimentation leverages the well-established AStock~\cite{zou2022astock}, a comprehensive benchmark for stock movement prediction based on Chinese news articles. Consistent with previous studies, we employ the universally recognized evaluation metrics: accuracy and F1 score. The dataset within AStock comprises 40,963 news articles and is pertinent to 3,680 Chinese market stocks, spanning from July 2018 to November 2021. 

Following the standard practice adopted in previous research~\cite{zou2022astock}, we adhere to the predetermined dataset splits, allocating 80\% of the data for model training, while the remaining 20\% is evenly split between validation and testing. Deviating from the traditional up/down movement prediction for stocks, \citet{zou2022astock} introduced a new prediction paradigm that attempts to forecast whether a stock mentioned in the news articles will outperform, remain neutral, or underperform compared to other stocks. We adopt this same prediction framework in our experimentation, providing a consistent basis for comparative evaluation.

\section{Experiment}

\subsection{Models}
We pre-finetune a variety of large pre-trained language models with our curated dataset:
BERT-Chinese~\cite{devlin-etal-2019-bert}, Multilingual-BERT~\cite{devlin-etal-2019-bert},Chinese-BERT~\cite{cui2021pre}, Mengzi-BERT~\cite{zhang2021mengzi}, Mengzi-BERT-Fin~\cite{zhang2021mengzi}.
The pre-finetuned models, referred to as IDED-BERT, IDED-mBERT, IDED-CBERT, IDED-Mengzi, and IDED-Mengzi-Fin, respectively, serve as the backbone of our study, offering a comprehensive exploration of different language modeling approaches in the context of stock movement prediction.

We draw on the state-of-the-art method Semantic Role Labeling Pooling (SRLP)~\cite{zou2022astock} as well as two renowned models for stock movement prediction: StockNet~\cite{xu-cohen-2018-stock} and HAN~\cite{hu2018listening}. Additionally, we include all standard pre-trained language models (PLMs) in our comparative analysis for a comprehensive assessment.

\subsection{Stock Movement Prediction}
The compiled experimental results are presented in Table~\ref{tab:Experimental results}.
Several key observations emerge from our experimental results. First, we noted that all pre-trained language models (PLMs) fell short of the performance achieved by SRLP, irrespective of the evaluation metric applied. This suggests that the more specialized SRLP model has certain advantages in this particular task compared to the more generalized PLMs.
Second, the techniques we have proposed demonstrate superior performance over the existing state-of-the-art in stock movement prediction. This superior performance was observed across all language models used, underscoring the value of our proposed pre-finetuning task. 

Third, within our proposed series of models, the IDED-Mengzi-Fin emerged as the best performer. This could be attributed to the fact that Mengzi-BERT-Fin is pre-trained with a corpus of financial documents, providing it with a domain-specific advantage. However, it's worth noting that the vanilla version of Mengzi-BERT-Fin was found to be the weakest performer among the pre-trained language models. These observations underscore the significance of impact duration as a crucial aspect for a deeper understanding of financial documents. Following pre-finetuning with our proposed task, Mengzi-Fin demonstrated a marked improvement, achieving the highest performance among all models evaluated.
Fourth, the stark contrast in performance between the vanilla PLMs and our IDED series models indicates that the vanilla models lack inherent capabilities for estimating impact duration. This validates our hypothesis that introducing the notion of impact duration can substantially enhance the performance of models.

\begin{table}[t]
  \centering
  \resizebox{0.9\columnwidth}{!}{
    \begin{tabular}{llrr}
    \Xhline{2\arrayrulewidth}
    \multicolumn{1}{c}{Type} & \multicolumn{1}{c}{Model}  & \multicolumn{1}{c}{Accuracy} & \multicolumn{1}{c}{F1} \\
    \hline
    \multirow{3}[2]{*}{SMP Model} & StockNet & 46.72\% & 44.44\% \\
          & HAN   & 57.35\% & 56.61\% \\
          & SRLP  & 61.76\% & 61.69\% \\
    \hline
    \multirow{5}[2]{*}{PLM} & BERT-Chinese & 58.36\% & 58.33\% \\
          & Multilingual-BERT & 58.36\% & 58.16\% \\
          & Chinese-BERT & 60.32\% & 60.33\% \\
          & Mengzi-BERT & 59.72\% & 59.70\% \\
          & Mengzi-BERT-Fin & 57.61\% & 57.44\% \\
    \hline
    \multirow{5}[1]{*}{Proposed} & IDED-BERT & \underline{63.17\%} & \underline{62.92\%} \\
          & IDED-mBERT & \underline{62.90\%} & \underline{62.84\%} \\
          & IDED-CBERT & \underline{61.68\%} & 61.36\% \\
          & IDED-Mengzi & \underline{64.05\%} & \underline{64.09\%} \\
          & IDED-Mengzi-Fin & \underline{\textbf{64.18\%}} & \underline{\textbf{64.15\%}} \\
    \Xhline{2\arrayrulewidth}
    \end{tabular}%
    }
    \caption{Experimental results. SMP and PLM denote stock movement prediction and pre-trained language model, respectively. \textbf{Bolded} results are the best performances, and \underline{Underlined} results are better than the state-of-the-art model, SRLP. }
  \label{tab:Experimental results}%
\end{table}%

\section{Discussion}
\subsection{Pre-Finetuning with Sentiment Analysis}

This subsection focuses on examining the effectiveness of the sentiment analysis task as a pre-finetuning stage for stock movement prediction. This exploration is paramount as it informs the choice of the optimal pre-finetuning task, in this case, being stock movement prediction.

As part of this investigation, we conduct sentiment annotations on the same set of posts employed in our Integrated Data Extraction and Derivation (IDED). The annotation process involves the expertise of two seasoned annotators. Following the protocols outlined in Section~\ref{sec:dataset}, the first expert evaluates all 8,760 posts while the second annotator examines a randomly selected sample of 1,000 posts. This allows for a calculation of inter-rater agreement.
The sentiment labels adopted in this process include bullish, bearish, and neutral, thereby encapsulating the range of potential investor sentiment. Notably, the Cohen-Kappa agreement rate stands at 62.21\%, indicative of substantial agreement between the two experts.

Upon the completion of the annotation process, we proceed to pre-finetune the five pre-selected language models using the newly minted sentiment labels. Adhering to the methodology described in Section~\ref{sec:dataset}, these finetuned models, hereafter referred to as Senti-BERT, Senti-mBERT, Senti-CBERT, Senti-Mengzi, and Senti-Mengzi-Fin, offer an additional point of comparison in our analysis.
In the spirit of comprehensive investigation, we introduce SKEP~\cite{tian-etal-2020-skep} into our comparative framework. SKEP is a pre-trained language model specifically designed for sentiment analysis, offering an interesting contrast to our models, which are initially trained on a variety of tasks.

The experimental results from these models are presented in Table~\ref{tab:Experimental results of pre-finetuning with sentiment labels}. A comparison between the PLM and Senti-PLM results shows an intriguing trend. Pre-training with sentiment analysis tasks appears to offer a minor enhancement in performance for stock movement prediction. Yet, the performance significantly degrades when the pre-finetuning stage employs sentiment analysis tasks, resulting in outcomes even inferior to the original pre-trained language models.
This unexpected observation underscores the significance of the judicious selection of pre-finetuning tasks. It illustrates that not all tasks using the same document corpus necessarily lead to an enhancement in model performance, cementing the vital role of task relevance in the pre-finetuning process.

\begin{table}[t]
  \centering
  \resizebox{\columnwidth}{!}{
    \begin{tabular}{llrr}
    \Xhline{2\arrayrulewidth}
    \multicolumn{1}{c}{Type} & \multicolumn{1}{c}{Model} & \multicolumn{1}{c}{Accuracy} & \multicolumn{1}{c}{F1} \\
    \hline
    \multirow{5}[2]{*}{PLM} & BERT-Chinese & 58.36\% & 58.33\% \\
          & Multilingual-BERT & 58.36\% & 58.16\% \\
          & Chinese-BERT & 60.32\% & 60.33\% \\
          & Mengzi-BERT & 59.72\% & 59.70\% \\
          & Mengzi-BERT-Fin & 57.61\% & 57.44\% \\
    \hline
    Senti-PLM & SKEP  & 60.66\% & 60.66\% \\
    \hline
    \multirow{5}[2]{*}{Sentiment PFT
} & Senti-BERT & 56.01\% & 55.88\% \\
          & Senti-mBERT & 55.97\% & 55.75\% \\
          & Senti-CBERT & 57.35\% & 56.97\% \\
          & Senti-Mengzi & 56.79\% & 56.12\% \\
          & Senti-Mengzi-Fin & 55.89\% & 55.34\% \\
    \hline
    \multirow{5}[1]{*}{Impact Duration PFT} & IDED-BERT & 63.17\% & 62.92\% \\
          & IDED-mBERT & 62.90\% & 62.84\% \\
          & IDED-CBERT & 61.68\% & 61.36\% \\
          & IDED-Mengzi & 64.05\% & 64.09\% \\
          & IDED-Mengzi-Fin & \textbf{64.18\%} & \textbf{64.15\%} \\
    \Xhline{2\arrayrulewidth}
    \end{tabular}%
    }
        \caption{Experimental results of pre-finetuning with sentiment labels. PFT denotes pre-finetune.}
  \label{tab:Experimental results of pre-finetuning with sentiment labels}%
\end{table}%

\subsection{Analysis of Impact Duration}
To further decipher the difference between long-term and short-term impact duration estimations, we employ the pointwise mutual information (PMI) technique. This aids in comparing word usage across the opinions designated with long-term and short-term impact duration estimations. PMI serves as a reliable tool for constructing sentiment dictionaries, as suggested by Khan et al.~\cite{khan2016sentimi}.
For this purpose, we compute the Impact Duration Estimation Difference (IDED) Score using the following equation:

\begin{equation}
\small
    \mathit{IDED Score}_w = \log_2 \frac{p(w,\mathit{LT})}{p(w)p(\mathit{LT})} - \log_2 \frac{p(w,\mathit{ST})}{p(w)p(\mathit{ST})},
\end{equation}
Here, $w$ symbolizes the word, and $LT$ and $ST$ denote whether $w$ is within the opinion labeled as ``Longer than 1 week'' and ``Within 1 week'', respectively.

Table~\ref{tab:Words with IDED Scores} features the top-ranked words within both groups. An intriguing pattern emerges here; investors typically assign longer impact duration estimations to fundamental events such as Cooperation, Land Acquisition/Sale, Integration, Structural Changes, and Business Expansion. In contrast, events associated with technical indicators generally attract shorter impact duration estimations. Examples include Price Increases, Trading Volume, and Monthly Moving Average, which tend to be estimated as ``Within 1 week''. Furthermore, trading actions such as Buy orders are often associated with short-term impacts.

\begin{table}[t]
  \centering
  \resizebox{\columnwidth}{!}{
    \begin{tabular}{lr|lr}
    \Xhline{2\arrayrulewidth}
    \multicolumn{2}{c|}{Longer than 1 week} & \multicolumn{2}{c}{Within 1 week} \\
    \multicolumn{1}{c}{Word} & \multicolumn{1}{c|}{IDED Score} & \multicolumn{1}{c}{Word} & \multicolumn{1}{c}{IDED Score} \\
    \hline
    Cooperation & 1.1839 & Raise & -3.6718 \\
    Land  & 1.1827 & Warrant & -3.6152 \\
    Integrate & 1.1761 & Trading Volume & -2.2548 \\
    Structure & 1.1380 & Monthly Moving Average & -2.2153 \\
    Expansion & 1.1111 & Buy   & -1.4592 \\
    \Xhline{2\arrayrulewidth}
    \end{tabular}%
    }
        \caption{Words with IDED Scores. }
  \label{tab:Words with IDED Scores}%
\end{table}%

\section{Conclusion}
This study initiates a novel research path that harnesses investors' estimations of impact duration, as part of a pre-finetuning task aimed at improving stock movement prediction. We circumvent the traditional method of direct news article annotation, instead opting for a more nuanced and practical approach that uses investor opinions. This allows us to capitalize on the expertise and intuitive predictions of investors, potentially adding a depth of insight unavailable through conventional methods. 
Our experiments underscore the efficacy of our approach, illustrating that the pre-finetuning with our proposed IDED consistently amplifies the performance of stock movement prediction across various language models. Importantly, we also discover the crucial role of task relevance in the pre-finetuning process, as integrating other tasks such as sentiment analysis into pre-finetuning does not guarantee a performance boost.

\section*{Limitations}
There are many languages, lots of financial instruments, and over a hundred markets in the world, but this paper only experiments on one language (Chinese), one financial instrument (stock), and one market (China). 
Although we cannot show the proposed method is a silver bullet for all markets, we think that this paper proposes a novel discussion for future work in both improving stock movement prediction and understanding investors' opinions. 
The results also show the usefulness of the proposed pre-finetuning task. 
Future works can examine whether the proposed research direction is also useful in other markets. 

Since the expert annotation is costly, the proposed dataset is mainly annotated by one expert. 
The other expert provides annotations for a subset of the dataset to check the quality. 
We think that 1,000 instances for checking the quality is a sizeable subset, and the agreement is also reached the substantial level~\cite{landis1977measurement}.
Additionally, instead of focusing on the performance of impact duration estimation, this paper uses the proposed dataset as the middle step for improving the performance of the stock movement prediction task. 
The evaluation of the benchmark stock movement prediction dataset also supports the usefulness of the proposed pre-finetuning task. 
Future works can extend our findings by enlarging the dataset or improving the annotation quality.

\bibliography{custom}

\appendix

\begin{table*}[t]
  \centering
  \resizebox{\textwidth}{!}{
    \begin{tabular}{ll}
    \hline
          & \multicolumn{1}{c}{URL} \\
    \hline
    \textbf{BERT-Chinese}~\cite{devlin-etal-2019-bert}  & \url{https://huggingface.co/bert-base-chinese} 
    \\
    \textbf{Multilingual-BERT}~\cite{devlin-etal-2019-bert} & \url{https://huggingface.co/bert-base-multilingual-cased}
    \\
    \textbf{Chinese-BERT}~\cite{cui2021pre} & \url{https://huggingface.co/hfl/chinese-bert-wwm}
    \\
     \textbf{Mengzi-BERT}~\cite{zhang2021mengzi} & \url{https://huggingface.co/Langboat/mengzi-bert-base} 
     \\
    \textbf{Mengzi-BERT-Fin}~\cite{zhang2021mengzi} & \url{https://huggingface.co/Langboat/mengzi-bert-base-fin} 
    \\

    \hline
    \end{tabular}%
    }
  \caption{Reference for the models in our experiments.}
  \label{tab:Reference for the models in our experiments}%
\end{table*}%

\section{Implement Detail}
\label{sec:Implement Detail}
In our experiment, we use the Hugging Face transformers package~\cite{wolf2019huggingface}. \footnote{\url{https://huggingface.co/docs/transformers/index}}
Intel Xeon Gold CPU and Nvidia Tesla V100 w/32GB are the CPU and GPU we used. 
Table~\ref{tab:Reference for the models in our experiments} provides the links to the models we used in the experiments.

Due to the size limitation of the submission system, we cannot upload the pretrained LMs to the system. 
We will release these LMs on the Hugging Face platform.
Please download them via the following anonymous link: \url{https://drive.google.com/drive/folders/1VhNt_vtuGVZm_0z_DInx_zli8I3NIOsR?usp=sharing}

\section{Dataset}
We will release our dataset for academic use under the CC BY-NC-SA 4.0 license.
It is worth noting that this social media platform has administrators, and the offensive posts were reported and removed manually.
We pay the annotators a salary that is higher than 20\% of the minimum wage in their country.
Please download it via the submission system.

\end{document}